\documentclass{article}

\usepackage{microtype}
\usepackage{graphicx}
\usepackage{subfigure}
\usepackage{booktabs} %

\usepackage{hyperref}
\usepackage{times}

\usepackage{amsmath,amsfonts,bm}

\def\eqref#1{equation~\ref{#1}}

\def\1{\bm{1}}

\DeclareMathAlphabet{\mathsfit}{\encodingdefault}{\sfdefault}{m}{sl}
\SetMathAlphabet{\mathsfit}{bold}{\encodingdefault}{\sfdefault}{bx}{n}

\newcommand{\R}{\mathbb{R}}

\usepackage[utf8]{inputenc} %
\usepackage[T1]{fontenc} %
\usepackage{hyperref} %
\usepackage{url} %
\usepackage{booktabs} %
\usepackage{amsfonts} %
\usepackage{nicefrac} %

\usepackage{stfloats}
\usepackage{rotating}
\usepackage{subfigure}
\usepackage[capitalise ]{cleveref}
\usepackage{amsmath}
\usepackage{amssymb}
\usepackage{mathtools}

\usepackage[accepted]{icml2020}

\icmltitlerunning{A simple defense against adversarial attacks on heatmap explanation}

\begin{document}

\twocolumn[
\icmltitle{A simple defense against adversarial attacks on heatmap explanations}

\icmlsetsymbol{equal}{*}
\begin{icmlauthorlist}
\icmlauthor{Laura Rieger}{dtu}
\icmlauthor{Lars Kai Hansen}{dtu}
\end{icmlauthorlist}

\icmlaffiliation{dtu}{DTU Compute, Technical University Denmark, 2800 Kgs. Lyngby, Denmark}

\icmlcorrespondingauthor{Laura Rieger}{lauri@dtu.dk}

\icmlkeywords{Machine Learning, neural networks, interpretability, XAI, adversarial attacks}

\vskip 0.3in
]
\printAffiliationsAndNotice{} 
\begin{abstract}
With machine learning models being used for more sensitive applications, we rely on interpretability methods to prove that no discriminating attributes were used for classification. A potential concern is the so-called "fair-washing"  - manipulating a model such that the features used in reality are hidden and more innocuous features are shown to be important instead. 

In our work we present an effective defence against such adversarial attacks on neural networks. By a simple aggregation of multiple explanation methods, the network becomes robust against manipulation. This holds even when the attacker has exact knowledge of the model weights and the explanation methods used. 

\end{abstract}

\section{Introduction}

In recent years machine learning algorithms have become more complex and are used for more important decisions. Since models, especially neural networks, are trained with large amounts of data, it is hard to oversee just what is hidden in the data and what correlations the model picks up on. 
Explainability methods present a solution for this \cite{hansen2019interpretability}. By looking at what features of the input were important for a classification, we can make sure that the classification is aligned with our  ethical convictions and understanding of the task.

\begin{figure}[hbt]
	\begin{center}
		\includegraphics[width=1\linewidth]{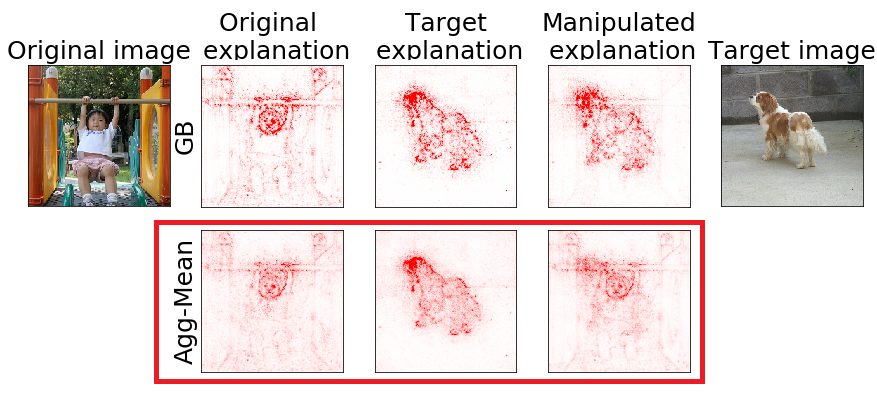}
	\end{center}
	\caption{Explanation methods (here Guided Backpropagation) are very vulnerable to  adversarial attacks. Our method, AGG-Mean, presents a simple but effective defence.
	}
	\label{fig:adv_attack_agg_short}
\end{figure}
It follows that there are many reasons why someone might want to manipulate an explanation, referred to as "fairwashing" \citep{aivodji2019fairwashing}. For example, a company may want to hide that they use discriminatory practices in their hiring  or someone may want to hide adversarial attacks on machine learning algorithms. Before explainability methods can be used and relied on in practice, we need to evaluate the the risk for this and find effective defences.
Previous works have shown that explainability methods are remarkably brittle against adversarial attacks. In practice, an attacker can effectively manipulate the explanation at will without any visual changes to the input that a human would pick up on \cite{dombrowski2019explanations,Ghorbani2019}.

We propose a simple way to ward against this potential security risk and make explainability methods more viable in deployment. Our approach is motivated by a key insight in machine learning: Ensemble models can reduce both bias and variance compared to applying a single model. A related approach was pursued for \textit{functional} visualization in neuroimaging \citep{hansen2001consensus}. 
 Based on this insight, we propose a way to aggregate explanation methods, \textit{AGG-Mean}. This approach is analysed theoretically and evaluated empirically.  In experiments on Imagenet, the aggregate is more robust to adversarial attacks than any single method. Even when the attacker has complete knowledge of the model weights and the explanation methods to be used as well as complete control over the input, the explanation stays robust as shown in \cref{fig:adv_attack_agg_short}.

\section{Related Work}
\label{sec:rel_work}
\subsection{Explanation methods}
The open problem of explainability is reflected in a lot of recent work \citep{Kindermans,Selvaraju,Bach2015,Zhang,Zhou2015,Ancona2017,Ribeiro,Rieger2018,Kim2017,lundberg2017unified,zintgraf2017visualizing, Simonyan2013, Zeiler2014,Selvaraju, Smilkov2017a, Sundararajan2017a,shrikumar2017learning,montavon2017explaining,chang2018explaining}. We focus on generating visual explanations for single samples. To our knowledge the first work in this direction was 
\citet{Simonyan2013} with \textit{Saliency Maps (SM)} that proposed backpropagating the output onto the input to gain an understanding of a neural network decision. The relevance for each input dimension is extracted by taking the gradient of the output w.\ r.\ t.\ to the input. This idea was extended by \citet{Springenberg2014} into \textit{Guided Backpropagation (GM)} by applying ReLU non-linearities after each layer during the backpropagation. Compared to Saliency, this removes visual noise in the explanation. 
\textit{Grad-CAM (GC)} from \citet{Selvaraju} is an explanation method, developed for use with convolutional neural networks. By backpropagating relevance through the dense layers and up-sampling the evidence for the convolutional part of the network, the method obtains coarse heatmaps that highlight relevant parts of the input image. 
\textit{Integrated Gradients (IG)} \citet{Sundararajan2017a} sums up the gradients from linearly interpolated pictures between a baseline, e.g.\ a black image, and the actual image.
\textit{SmoothGrad (SG)} filters out noise from a basic saliency map by creating many samples of the original input with Gaussian noise \citep{Smilkov2017a}. The final saliency map is the average over all samples. 
In concurrent work, \citep{bhatt2020evaluating} also proposed aggregating explanation methods, albeit with the goal of decreasing complexity rather than vulnerability. Finally, \citep{yeh2019fidelity} showed that a combination of  two popular  explanation methods is optimal in terms of fidelity.

\subsection{Adversarial attacks on explanation methods}
While adversarial examples for classification are well-known, recently there has been growing interest in adversarial manipulation of explanations \citep{Ghorbani2019,heo2019fooling,dombrowski2019explanations}.
Attacks on explanation can serve multiple purposes including "fairwashing" \cite{aivodji2019fairwashing}. All of these methods exploit the fully differentiable nature of neural networks and iteratively update the input (or the model weights) to change the explanation while only minimally changing the input and output. The goal is to manipulate the explanation while keeping the input and output (visually) similar. It is assumed that the network architecture and weights are known and that either the input (\citep{Ghorbani2019,zhang2018interpretable,dombrowski2019explanations}) or the network weights \cite{heo2019fooling} can be changed by the attacker.

Focussing on changing the input,
 \citet{Ghorbani2019},\citet{zhang2018interpretable} and \citet{dombrowski2019explanations} attack the explanation by manipulating the image, not changing the network weights. Interestingly, \citet{zhang2018interpretable} discuss the transferability of attacks and conclude that attacks are not that transferable. If the attacker is allowed to modify networks weights, as in \citet{heo2019fooling}, the attacks generalize to all the considered explanation methods.
 We are interested in the more realistic situation where the attacker can modify the input but not the network. We investigate the transferability, c.f., \citet{zhang2018interpretable}, and hypothesise that the limited transferability leads to improved robustness of the ensemble explanation.
 While ensemble methods have been proposed earlier as a defense for attacks on the label \cite{tramer2017ensemble}, they have not previously been investigated as a defense mechanism against attacks on explanations.
 
\section{Averaging explanation methods to reduce vulnerability}

\label{sec:agg}

\subsection{Averaging explanations}
All currently available explanation methods have weaknesses that are inherent to the approach and include significant uncertainty in the resulting heatmap \citep{Kindermans, Adebayo, Smilkov2017a}.
A natural way to mitigate this issue and reduce noise is to combine multiple explanation methods. Ensemble methods have been used for a long time to reduce the variance and bias of machine learning models. We apply the same idea to explanation methods and build an ensemble of explanation methods. Ensemble methods have also been previously used to defend against adversarial attacks on neural network outputs \cite{pang2019improving,liao2018defense}, motivating the usage of an explanation ensemble to defend against attacks on the explanation. 

We assume a neural network $ F: X \mapsto y $ with $ X \in \R^{m \times m} $ and a set of explanation methods $ \{e_j\}_{j=1}^{J} $ with $ e_j : X,y,F \mapsto E$ with $ E \in \R^{m \times m} $. We write $ E_{j,n} $ for the explanation obtained for $ X_n $ with method $ e_j $ and denote the mean aggregate explanation as $ \bar{e} $ with $\bar{E}_n = \frac{1}{J}\sum_{j=1}^J E_{j,n}$. While we assume the input to be an image $ \in \R^{m \times m} $, this method is generalizable to inputs of other dimensionalities as well.

To get a theoretical understanding of the aggregation, we hypothesize the existence of a 'true' explanation $ \hat{E} _n$. This allows us to quantify the error of an explanation method as the mean squared difference between the 'true' explanation and an explanation procured by an explanation method, i.e.\ the MSE. 

For clarity we subsequently omit the notation for the neural network. We write the error of explanation method $ j $ on image $X_n$ as
$ \text{err}_{j,n}= || {E}_{j,n} - \hat{E}_n||^2 $
with \[ \text{MSE}(E_j) = \frac{1}{N} \sum_{n} \text{err}_{j,n} \]
and $ \text{MSE}(\bar{E})= \frac{1}{N}\sum_{n} || {\bar{E}}_{n} - \hat{E}_n||^2$ is the MSE of the aggregate. 
The typical error of an explanation method is the mean error over all explanation methods \[ \overline{\text{MSE}} =\frac{1}{J}\sum_j \text{MSE}(E_j). \]
With these definitions we can do a standard bias-variance decomposition \citep{geman1992neural}. Accordingly we can show the error of the aggregate will be less that the typical error of explanation methods,
\begin{eqnarray} 
\overline{\text{MSE}} = &\frac{1}{N}\sum_{n} \frac{1}{J}\sum_{j}||\hat{E}_n - E_{j,n} ||^2 \\
= &\frac{1}{N}\sum_{n} ||\hat{E}_n - \bar{E}_n ||^2 \\
&+ \frac{1}{NJ}\sum_{n,j}|| \bar{E}_n - E_{j,n} ||^2 ,\nonumber
\end{eqnarray}
hence,
\begin{eqnarray}
\label{eq:specific_methods}
\overline{\text{MSE}} = &\frac{1}{J}\sum_{j} \frac{1}{N}\sum_{n} ||\bar{E}_n - E_{j,n}||^2 + \text{MSE}(\bar{E})\\
 \geq&\hspace{-45mm}\text{MSE}(\bar{E}). \nonumber
\end{eqnarray}
A detailed calculation is given in the appendix.
The error of the aggregate $\text{MSE}(\bar{E}) $ is less than the typical error of the participating methods. The difference - a `variance' term - represents the epistemic uncertainty and only vanishes if all methods produce identical maps.
By taking the average over all available explanation methods, we reduce the variance of the explanation compared to using a single method. To obtain this average, we normalize all input heatmaps such that the relevance over all pixels sum up to one. This reflects our initial assumption that all individual explanation methods are equally good estimators. We refer to this approach as \textit{{AGG-Mean}}.
\begin{equation} \label{eq:aggmean}
E_{\text{AGG-Mean},n} =\frac{1}{J} \sum_{j=1}^{J} E_{j,n} 
 \end{equation}

\subsection{Adversarial scenarios}
\label{ssub:attack}
With the increasing interest and practical importance of explainability of neural networks the interest in methods for manipulation and control of explanations is also increasing. A typical scenario is to make imperceptible changes to the input of the neural network such that the output/label is unchanged while the explanation changes according to a given goal. Such effort could, e.g., be used to hide bias or other fairness issues a given classifier might have. 

\citet{dombrowski2019explanations} showed that explanations can be made more robust by replacing the ReLU nonlinearity with a Softplus function. However, this requires changing the network and using a different architecture for classification and explanation, which is highly undesirable as it defeats the purpose of the explanation. The analysis of \citet{zhang2018interpretable} showed that transferability of attacks is limited, hence, our ensemble of multiple explanations may offer robustness also towards certain types of adversarials. 

In the following we will assume an attacker who has full knowledge of the neural network, including the architecture and weights. In contrast to \citet{heo2019fooling}, however, we will assume that the attacker cannot \emph{change} the neural network, following \citet{dombrowski2019explanations,Ghorbani2019}. Furthermore the attacker has full control over the input to the neural network. The goal is to adversarially manipulate the image according to a predefined objective. 

In the following, $ x $ will refer to the original image. $ \hat{x} $ is the 'target' image. The objective is to produce an adversarial input $ x' $ with $ x' \approx x $ but the explanation $ E_{x'} \approx E_{\hat{x}}$. While we focus on assimilating the explanation map of another input as in \citet{dombrowski2019explanations}, all techniques introduced can be readily adapted to other objectives, f.e.\ to move the mass center of the explanation. 

Exploring the robustness of aggregates of multiple explanation methods we concentrate on the following two scenarios:
\paragraph{Arsenal of explanation methods}
In this scenario we have a pool of potential explanation methods. The attacker does not have knowledge of what explanation method is used and optimizes for a different explanation method than is used by the defender. 
The success of the attack depends on how readily an attack of one explanation method translates to another method. 

We hypothesize that attacks on explanation methods are fragile and do not translate well across explanation methods, as they exploit locally high variances in the gradient landscape. This hypothesis is examined in \cref{ssub:attack}.
\paragraph{Aggregation of explanation methods}
In this more challenging scenario we aggregate multiple explanation methods as described in \cref{eq:aggmean}. The attacker knows the exact explanation methods and ratio going into the mixture and attacks this aggregation.

Many attribution-based methods are utilizing the gradient $ \frac{\delta y}{\delta x}(x) $ of the output to create an explanation. Due to the non-linearity of the neural network, the gradient can change rapidly with small distances in input space \cite{dombrowski2019explanations,Ghorbani2019}. Attacks on explanation methods exploit this vulnerability. 

\section{Experiments}
\label{sec:experiments}

We evaluate how robust aggregated methods are against adversarial attacks, compared to unaggregated methods. In all cases we assume that the attacker has full knowledge of the network architecture and weights (white box attack) but cannot change them. However, the attacker has full control over the input. 

Following \cite{dombrowski2019explanations} we run experiments on a pretrained VGG16\footnote{Models retrieved from \href{https://github.com/keras-team/keras}{{https://github.com/keras-team/keras}}.} 
We consider Layerwise Relevance Propagation (LRP), Saliency Mapping (SM), Guided Backprop (GB) and Integrated Gradients (IG) as explanation methods. The latter was not used in the aggregation.
 
 Unless otherwise noted we followed \citep{dombrowski2019explanations} in the choice of hyperparameters for attacking explanation methods. 
In the appendix we show that our defence also works against the attack as proposed in \citet{Ghorbani2019}.
 
 Since the ReLU function used in neural networks is not twice differentiable, we replace it with a differentiable approximation, SoftPlus for the iterative creation of the adversarial input. The final manipulated heatmaps are created with the ReLU non-linearity. Further details about the experiments are in the appendix.

We consider the two scenarios introduced in \cref{ssub:attack}. In all cases, the objective of the attacker is to make the explanation of input $ E_{x'} \approx E_{\hat{x}} $ while keeping $ x' \approx x $. To do this, the attacker manipulates $ x' $.

We visually confirmed that the adversarial images look similar to the input images and provide examples in the appendix. 
We measure the difference between the start explanation $ E_x $, target explanation $ E_{\hat{x}} $ and adversarial explanation $ E_{x'} $ with the MSE (Mean Square Error), the PCC (Pearson Correlation Coefficient) and the top-\textit{k} intersection with k being ad-hoc set to 10\% \citep{dombrowski2019explanations,Ghorbani2019}. 

 In all metrics, explanations obtained with different methods have different 'base' values (similarity between the explanations of two randomly chosen images) due to structural differences between explanation methods. To account for this, we consider for each similarity metric $ m_{\text{sim}} $ the difference $ m_{\text{sim}}(E_{\hat{x}}, E_{x'}) - m_{\text{sim}}(E_{\hat{x}}, E_x)$, i.e.\ how much \textit{more} similar the attack makes $ E_{x'} $ look to $ E_{\hat{x}} $. For the MSE, this results in a negative score, since the difference between the target and the attack is less than between the target and the starting point. For all metrics, the ideal score is $ 0 $, i.e.\ the attack did not change the explanation at all. Thus, for MSE a high value is desirable, for PCC and top-\textit{k} union a low value is desirable.

\paragraph{Transferability of attacks on explanation methods}
\begin{figure}[!ht]
	\begin{center}
		\includegraphics[width=\linewidth]{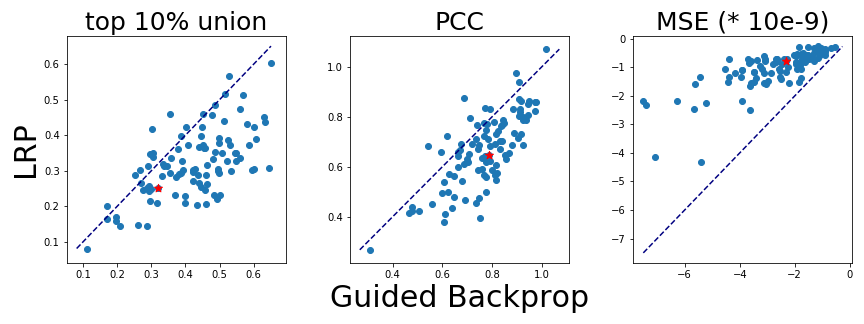}
	\end{center}
	\caption{ Attacking one method does not translate to attacks on the other methods.  Similarity metrics (topK and PCC) should be low, MSE should be high. Since for most samples \textit{topK} and \textit{PCC} are higher for the attacked method (Guided Backprop) than for LRP, attacks on Guided Backprop do not translate well to LRP. Red dot is the single sample visualized in \cref{fig:attack_example_pumpkin} }
	\label{fig:adv_attack}
\end{figure}

\begin{figure*}[tb]
	\begin{center}
		\includegraphics[width=.6\linewidth]{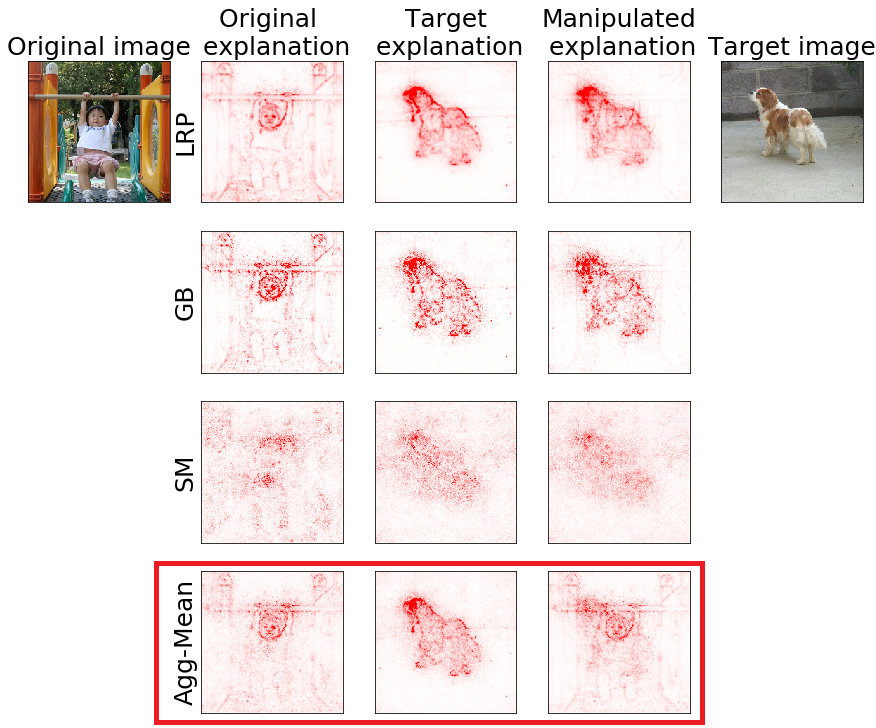}
	\end{center}
	\caption{Attacking explanation methods. Each method is targeted individually. AGG-Mean visibly preserves original explanation best, thus being the most resistant to adversarial attacks. To visualize details better we clipped values at the 99 percentile.  We provide the adversarial input images in the appendix. 
	}
	\label{fig:adv_attack_agg}
\end{figure*}

Visually comparing the success of an attack on AGG-Mean on the y-axis compared to unaggregated method (Guided Backprop) on the x-axis. Similarity metrics (\textit{topK} and \textit{PCC}) should be low, MSE should be high for less similarity between target and adversarial. Since for most samples \textit{topK} and \textit{PCC} are lower for AGG-Mean than for Guided Backprop, AGG-Mean is more robust than Guided Backprop. The red dot is the sample visualized in \cref{fig:adv_attack_agg}.

The lack of transferability results are in line with the findings of \citep{zhang2018interpretable}. 

\begin{figure}[!thb]
	\begin{center}
	\includegraphics[width=\linewidth]{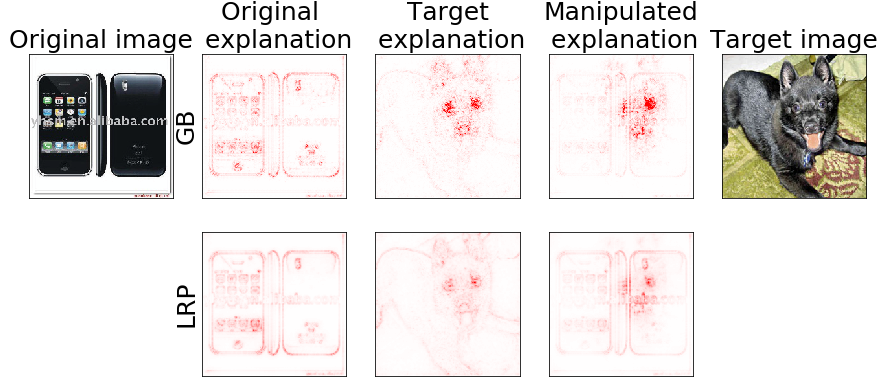}
	\end{center}
	\caption{Adversarial attacks do not transfer well between explanation methods. The adversarial input was calculated to attack GB (upper row). We then extracted the explanation with LRP (lower row). The attack does not transfer well to LRP. 
		To visualize details better we clipped values at the 99th percentile. }
	\label{fig:attack_example_pumpkin}
\end{figure}
We consider a case where the attacker does not know what explanation method is used, i.e.\ we attack a different explanation method than the one that is used. This would be the case if the defender has not made the specific explanation method used public or is choosing one at random to ward off attacks.
If the attack translates well, i.e.\ the image manipulation fools both methods, similarity metrics should be similar for both explanation methods. 

In \cref{fig:adv_attack} we provide results for attacking Guided Backpropagation and extracting an explanation with LRP. For a hundred samples we visualize for each sample the respective similarity metrics for both explanation methods in \cref{fig:adv_attack}. If the attack translates well, the points should lie on the identity line in \cref{fig:adv_attack}. Samples below the identity line for PCC and topK and above for MSE indicate that the attack does not generalize to other explanation methods.

As visible in \cref{fig:adv_attack} and anecdotally in \cref{fig:attack_example_pumpkin} (red data point in \cref{fig:adv_attack}), attacks perform much worse on other methods (here LRP) than the targeted one (here GB).  We provide statistics for other combinations in the appendix. 

\paragraph{Attacking aggregations of explanation methods}
\label{subsec:attack_eval}

In the second scenario the attacker knows that the explanations are aggregated and attacks the aggregation. We aggregate LRP, GB and SM and compare against those methods as well as Integrated Gradient. IG was not included in the aggregation as it requires sampling for each step, making it computationally much more expensive than the other methods. 
\begin{figure}[H]
	\begin{center}
		\includegraphics[width=\linewidth]{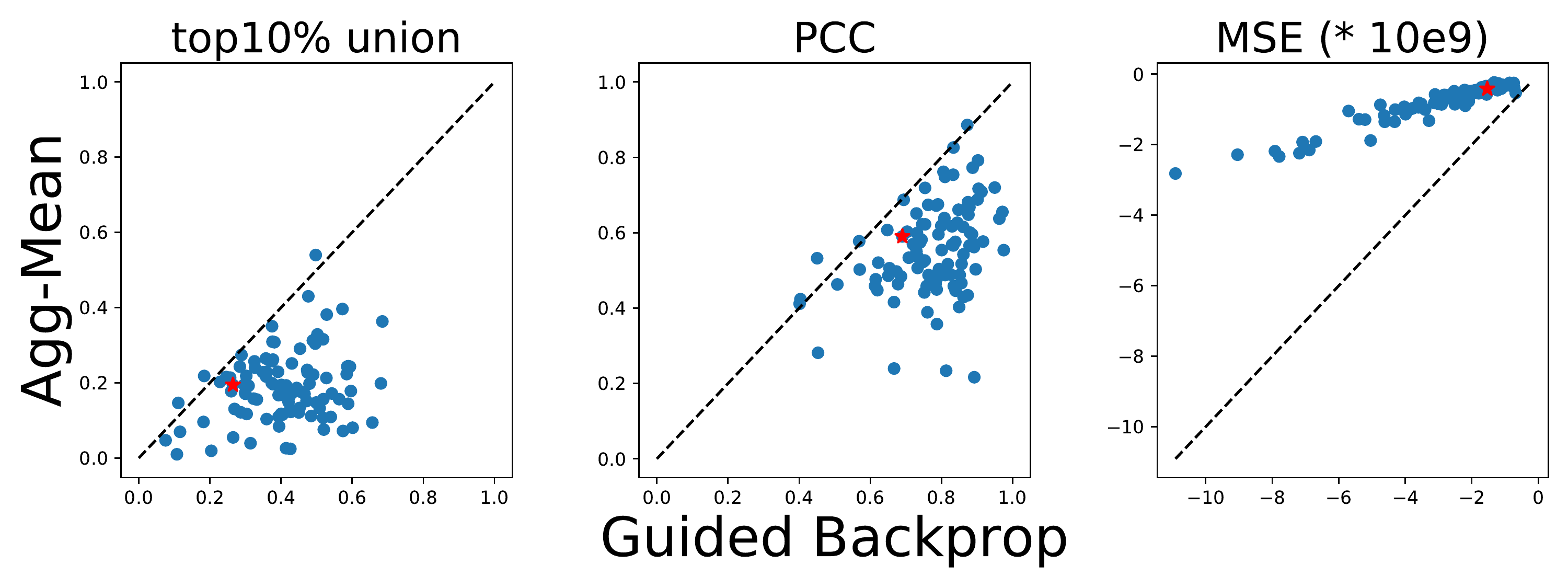}
	\end{center}
	\caption{Visually comparing the success of an attack on AGG-Mean (y-axis)  to the unaggregated method Guided Backprop (x-axis). Similarity metrics (\textit{topK} and \textit{PCC}) should be low, MSE should be high for less similarity between target and adversarial. Since \textit{topK} and \textit{PCC} are lower for AGG-Mean than for Guided Backprop, AGG-Mean is more robust than Guided Backprop. Red dot is sample visualized in \cref{fig:adv_attack_agg}. }
	\label{fig:adv_attack_agg_summary}
\end{figure}

In \cref{tab:results_attacks} we provide  metrics averaged over a hundred samples.  AGG-Mean outperforms unaggregated methods. We also provide a direct comparison to GuidedBackprop in \cref{fig:adv_attack_agg_summary}. To give an intuition on what differences in the  metrics  look like, we visualize a sample (red dot in \cref{fig:adv_attack_agg_summary})  in \cref{fig:adv_attack_agg}. We see that AGG-Mean opposed to the unaggregated methods largely preserves the original heatmap after the attack. More examples are provided in the appendix.

The resilience of the aggregate to attacks can be understood in terms of averaging induced smoothness. In \citep{dombrowski2019explanations}
the beneficial effects of averaging in the SmoothGrad method are described. As noted in \citep{dombrowski2019explanations} SmoothGrad is computationally expensive.
We conjecture that the diversity of the methods involved in the present aggregate implies that smoothing can be achieved at less computational effort.

\begin{table}[hbt]
	\caption{Evaluation scores across methods and architectures on a hundred samples. \textit{AGG-Mean} surpasses all considered methods in all metrics. All SE $\leq 0.02$. }
	\label{tab:results_attacks}
	\vskip 0.15in
	\begin{center}
		\begin{small}
			\begin{sc}
				\begin{tabular}{lrrr}
					\toprule
					{} & MSE (*10e-9) & PCC & top 10\% Union \\
					method & & & \\
					\midrule
					SM 		        & -0.92 	& 0.74  	& 0.40  \\
					GB 				& -3.25		& 0.77 		& 0.42 \\
					LRP             & -1.45		& 0.81  	& 0.49  \\
					IG 				& -1.76		& 0.82 			& 0.47\\
					AGG-Mean 	& \textbf{-0.89}  	& \textbf{0.54}	& \textbf{0.24} \\
					\bottomrule
				\end{tabular}
				
			\end{sc}
		\end{small}
	\end{center}
	\vskip -0.15in
\end{table}

\section{Conclusion}
In recent times, attacks on explanation method have received increased attention as the so-called "fairwashing", manipulating explanations to more innocuous ones, has become a concern.

We provide a simple and intuitive approach to defend against such attacks that does not require the model to be changed in any way and is computationally inexpensive. This approach is explored theoretically. We then provided experimental evidence that aggregations are a more robust to adversarial manipulation than individual explanation methods. 

Perhaps surprisingly, a simple average with the originally attacked method included induces a more robust explanation than replacing the explanation method with a different one. 
In \cite{dombrowski2019explanations} arguments are presented that the observed vulnerability is due to non-smoothness of contemporary networks. It is also argued that averaging as in SmoothGrad increases robustness. We theorize that the averaging of the diverse set of explanation methods involved in the aggregate creates similar smoothness. We noted that in contrast to \cite{dombrowski2019explanations}, the aggregate does not require modification (smoothing) of the network. 

We hope that our approach will be useful to make neural networks more transparent and increase their credibility as they are  applied in real-life scenarios.

\clearpage
\bibliography{bib}
\bibliographystyle{icml2020}

\clearpage

\appendix
\onecolumn
\section{Appendix}
\subsection{Aggregating explanation methods to reduce variance - detailed derivation}
\label{sec:agg_appendix}
All currently available explanation methods have weaknesses that are inherent to the approach and include significant noise in the heatmap \citep{Kindermans, Adebayo, Smilkov2017a}.
A natural way to mitigate this issue and reduce noise is to combine multiple explanation methods. Ensemble methods have been used for a long time to reduce the variance and bias of machine learning models. We apply the same idea to explanation methods and build an ensemble of explanation methods.

We assume a neural network $ F: X \mapsto y $ with $ X \in \R^{m x m} $ and a set of explanation methods $ \{e_j\}_{j=1}^{J} $ with $ e_j : X,y,F \mapsto E$ with $ E \in \R^{mxm} $. We write $ E_{j,n} $ for the explanation obtained for $ X_n $ with method $ e_j $ and denote the mean aggregate explanation as $ \bar{e} $ with $\bar{E}_n = \frac{1}{J}\sum_{j=1}^J E_{j,n}$. While we assume the input to be an image $ \in R^{mxm} $, this method is generalizable to inputs of other dimensions as well.

We define the error of an explanation method as the mean squared difference between a hypothetical 'true' explanation and an explanation procured by the explanation method, i.e.\ the MSE. For this definition we assume the existence of the hypothetical 'true' explanation $ \hat{E} _n$ for image $ X_n $.

For clarity we subsequently omit the notation for the neural network.

We write the error of explanation method $ j $ on image $X_n$ as
$ err_{j,n}= || {E}_{j,n} - \hat{E}_n||^2 $
with \[ \text{MSE}(E_j) = \frac{1}{N} \sum_{n} err_{j,n} \]
and $ \text{MSE}(\bar{E})= \frac{1}{N}\sum_{n} || {\bar{E}}_{n} - \hat{E}_n||^2$ is the MSE of the aggregate. 
The typical error of an explanation method is represented by the mean

\begin{equation} 
\begin{split}
\overline{\text{MSE}} &= \frac{1}{N}\sum_{n} \frac{1}{J}\sum_{j}||\hat{E}_n - E_{j,n} ||^2 \\
&= \frac{1}{NJ}\sum_{n,j}||\hat{E}_n - E_{j,n} + \bar{E}_n - \bar{E}_n||^2 \\
&= \frac{1}{NJ}\sum_{n,j}||(\hat{E}_n - \bar{E}_n) +( \bar{E}_n - E_{j,n} ) ||^2 \\
&= \frac{1}{NJ}\sum_{n,j}|| \hat{E}_n - \bar{E}_n ||^2 + || \bar{E}_n - E_{j,n} ||^2 +\frac{1}{NJ}\sum_{n,j}\left( 2{\rm Tr}\left[(\hat{E}_n - \bar{E}_n)(\bar{E}_n - E_{j,n} )\right]\right) \\
&= \frac{1}{N}\sum_{n} ||\hat{E}_n - \bar{E}_n||^2 + \frac{1}{NJ}\sum_{n,j}|| \bar{E}_n - E_{j,n} ||^2 + 2\frac{1}{N}\sum_{n} {\rm Tr}\left[(\hat{E}_n - \bar{E}_n)\left( \frac{1}{J}\sum_{j} (\bar{E}_n - E_{j,n} )\right)\right] \\
&= \frac{1}{N}\sum_{n} ||\hat{E}_n - \bar{E}_n ||^2 + \frac{1}{NJ}\sum_{n,j}|| \bar{E}_n - E_{j,n} ||^2 + 2\frac{1}{N}\sum_{n}{\rm Tr}\left[ (\hat{E}_n - \bar{E}_n)\underbrace{\frac{1}{J}\sum_{j} (\bar{E}_n - E_{j,n} ) }_{=0} \right]\\
&= \frac{1}{N}\sum_{n} ||\hat{E}_n - \bar{E}_n ||^2 + \frac{1}{NJ}\sum_{n,j}|| \bar{E}_n - E_{j,n} ||^2 ,\nonumber
\end{split}
\end{equation}

hence,
\begin{align}
\label{eq:specific_methods2}
\overline{\text{MSE}} &=\text{MSE}(\bar{E}) + \frac{1}{NJ}\sum_{n,j} \underbrace{||\bar{E}_n - E_{j,n}||^2}_{\text{epistemic uncertainty}} \geq \text{MSE}(\bar{E}) \nonumber
\end{align}
The error of the aggregate $\text{MSE}(\bar{E}) $ is less than the typical error of the participating methods. The difference - a `variance' term - represents the epistemic uncertainty and only vanishes if all methods produce identical maps.

\subsection{Experimental setup }
\subsubsection{General}

For AGG-Var, we add a constant to the denominator. We set this constant to $ 10 $ times the mean std, a value chosen empirically after trying values in the range of $ [1, 10, 100] $ times the mean. \\
Evaluations were run with a set random seed for reproducibility. SE were reported either for each individual result or if they were non-significant in the caption to avoid cluttering the results. \\ 
All experiments were done on a Titan X.

\subsubsection{ImageNet}

We downloaded the data from the ImageNet Large Scale Visual Recognition Challenge website and used the validation set only. No images were excluded. The images were preprocessed to be within $ [-1,1] $ unless a custom range was used for training (indicated by the preprocess function of keras).

\subsubsection{Details about attacking explanation methods}

For a range of explanation methods we chose to compare against LRP, Gradient, Guided Backpropagation and Integrated Gradients, a range of well-known and well-established explanation methods \cite{Sundararajan2017a,Bach2015,Springenberg2014,Simonyan2013}. Since Integrated Gradients is thirty times more computationally expensive than other methods, we did not include it in the aggregation as it would have slowed down experiments considerably. 

Unless otherwise noted, all metrics are computed as the average of a hundred data samples with mean and SE. Informally, we also found that the MSE does not align well with perceived changes in the explanations, likely due to it being susceptible to outliers. 

We used a pretrained VGG16 for all experiments attacking explanation methods \cite{simonyan2014very}. 

\clearpage

\subsection{Alignment between human attribution and explanation methods}
\label{sec:human_eval}
We want to quantify whether an explanation method agrees with human judgement on which parts of an image should be important. While human annotation is expensive, there exists a benchmark for human evaluation introduced in \cite{Mohseni2018}. 
\begin{figure*}[ht]
	\begin{center}
		\includegraphics[width=\linewidth]{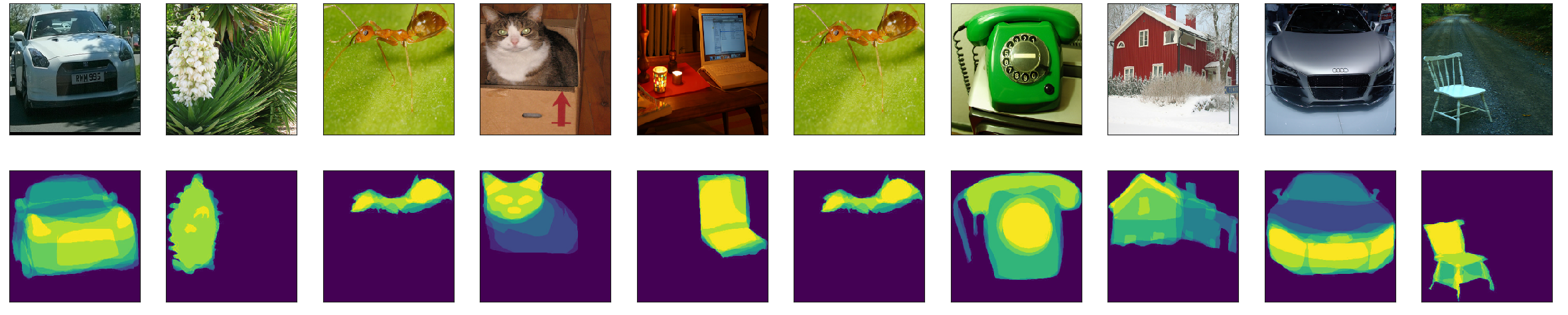}
	\end{center}
	\caption{Example images from \cite{Mohseni2018} with human-annotated overlays.}
	\label{fig:sample_human}
\end{figure*}
The benchmark includes ninety images of categories in the ImageNet Challenge (ten images were excluded due to the category not being in the ImageNet challenge) and provides annotations of relevant segments that ten human test objects found important. Example images are shown in \cref{fig:sample_human}.

While human evaluation is not a precise measure, we still expect some correlation between neural network and human judgement. 

To test the alignment, we calculate the cosine similarity,
\[\text{similarity}(e_j) = \frac{\sum_{n=1}^{N} A_n E_{j,n}}{\sqrt{\sum_{n=1}^{N}A_n^2}\sqrt{\sum_{n=1}^{N}E_{j,n}^2}}\]

between the human annotation and the explanations produced by the respective explanation methods. $ A_n $ is the human annotation of what is important for image $ X_n $

Since the images in this dataset are 224x224 pixel large, we only compute the cosine similarity for the network architectures where pretrained networks with this input size were available. 

We see that \textit{AGG-Mean} and \textit{AGG-Var} perform on-par with the best methods (SmoothGrad and GradCAM). While the aggregated methods perform better than the average explanation method, they do not surpass the best method. 

When we combine the two best-performing single methods, SmoothGrad and GradCAM, we surpass each individual method. We hypothesize that this is because the epistemic uncertainty is reduced by the aggregate.

\begin{table*}[!h]
	\caption{Cosine similarity between heatmap and human annotated benchmark. All SE below $ 0.05 $}
	\vskip 0.15in
	\begin{center}
		\begin{small}
			\begin{sc}

\begin{tabular}{lrrr}
\toprule
	{} &  ResNet101 &  ResNet50 &  VGG19 \\
	Method      &            &           &        \\
	\midrule
	AGG-Mean    &       0.63 &      0.66 &   0.64 \\
	AGG-Var     &       0.66 &      0.68 &   0.67 \\
	GB          &       0.42 &      0.49 &   0.47 \\
	GC          &       0.60 &      0.62 &   0.60 \\
	IG          &       0.45 &      0.45 &   0.47 \\
	Mean(SG+GC) &       \textbf{0.69} &      \textbf{0.70} &   \textbf{0.65} \\
	SG          &       0.63 &      0.65 &   0.59 \\
	SM          &       0.45 &      0.45 &   0.47 \\
\bottomrule
\end{tabular}

			\end{sc}
		\end{small}
	\end{center}
	\vskip -0.15in
\end{table*}
\clearpage

\subsection{Details about attacking explanation methods}
\paragraph{Choice of explanation methods}
We focused on explanation methods that have previously been shown to be susceptible to adversarial attacks. As such, we did not include GradCAM in the experiments, neither as a comparison or in the aggregation. 

Different explanation methods have different computational loads. Notably, SmoothGrad and IntegratedGradients involve the sampling of many explanations for a single pass, increasing computation times by the number of samples ()  and were not included in the aggregation but as a comparison. 
\paragraph{Choice of hyperparameters}
We  followed \cite{dombrowski2019explanations} for the choice of hyperparameters in learning rate and beta growth. For AGG-Mean we chose a learning rate of $ 10^{-3} $ and $ 1500 $ iterations for the attack.

hyperparameter choice when attacking explanation methods, using a learning rate of $ 10^{-3} $. 

\subsubsection{More examples}
\begin{figure*}[hbt]
	\begin{center}
		\includegraphics[width=\linewidth]{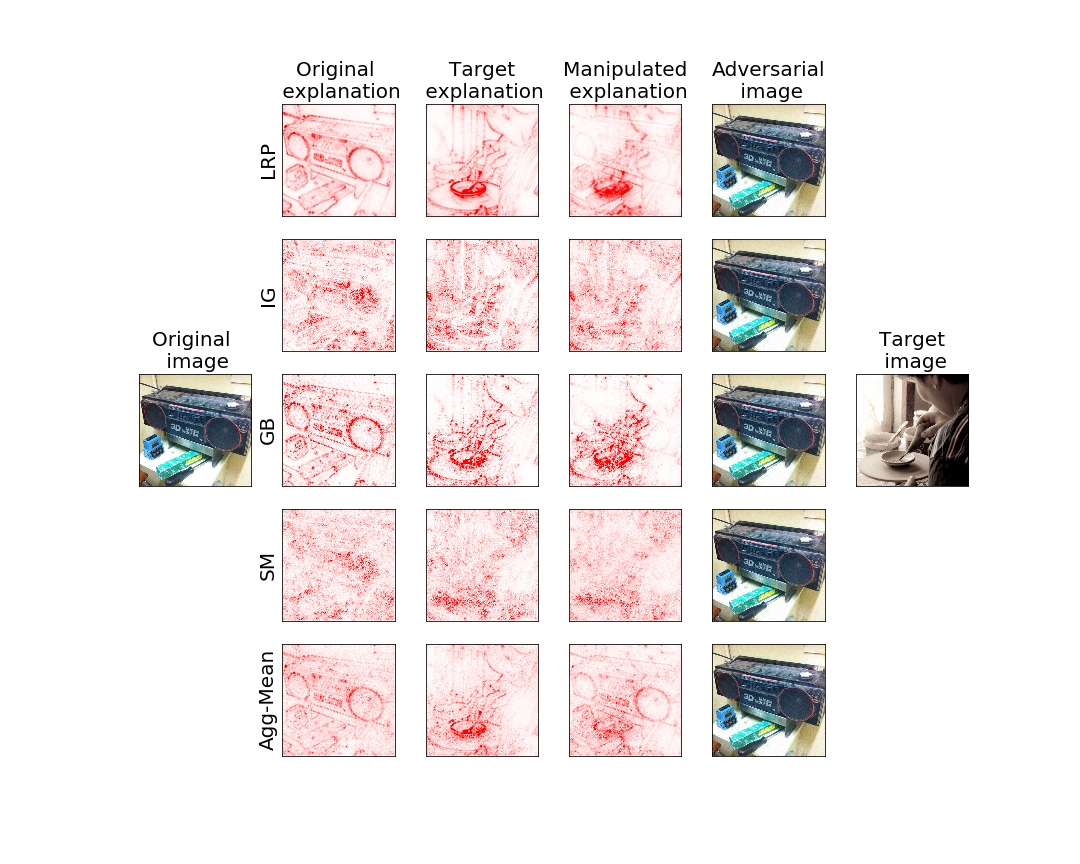}
	\end{center}
	\caption{Attack shown in the main text, including the adversarial input images. There are no visual differences for any of the adversarial inputs.}
	\label{fig:attack0}
\end{figure*}

\begin{figure*}[hbt]
	\begin{center}
		\includegraphics[width=\linewidth]{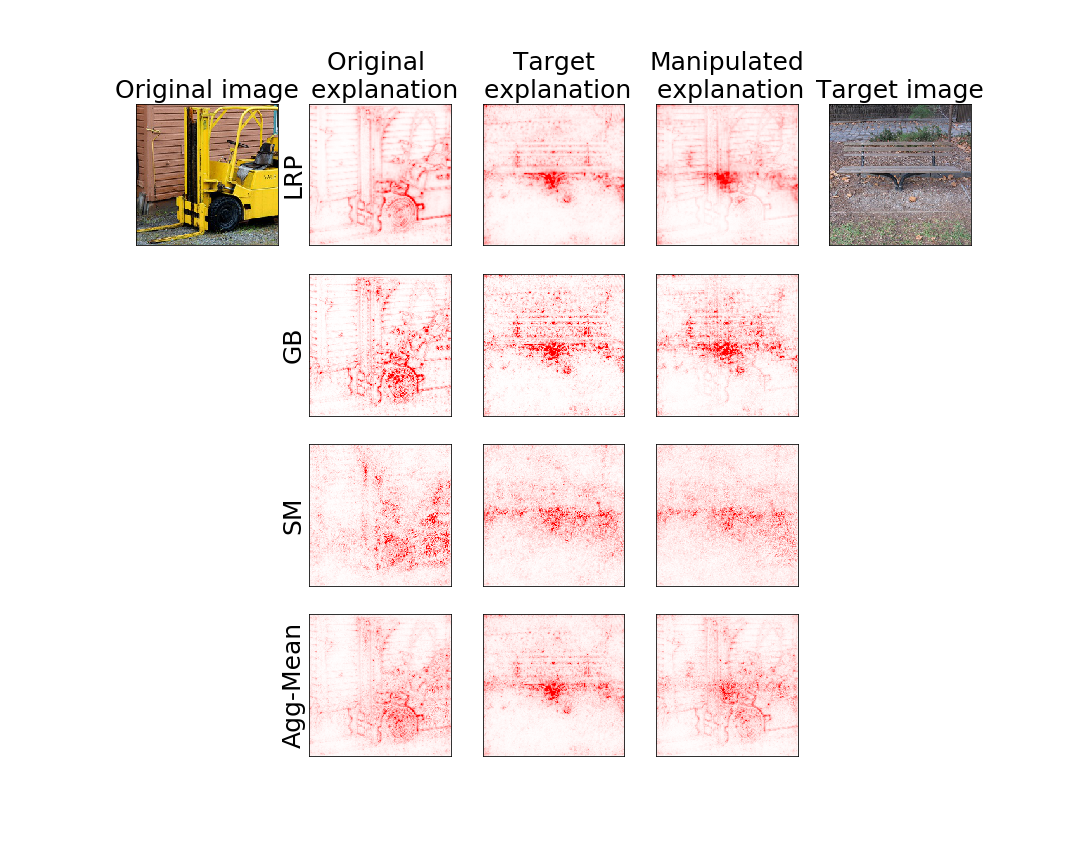}
	\end{center}
	\caption{Appendix example 1}
	\label{fig:attack1}
\end{figure*}

\begin{figure*}[hbt]
	\begin{center}
		\includegraphics[width=\linewidth]{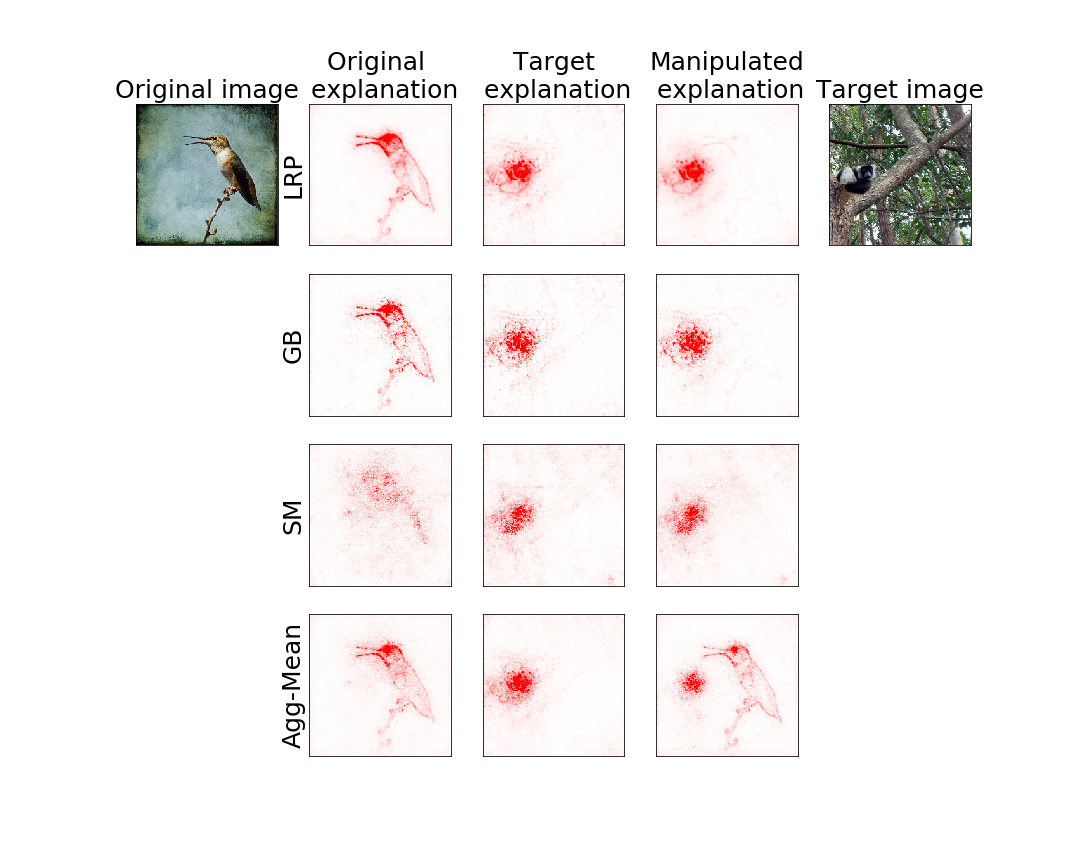}
	\end{center}
	\caption{Appendix example 2}
	\label{fig:attack3}
\end{figure*}

\begin{figure*}[hbt]
	\begin{center}
		\includegraphics[width=\linewidth]{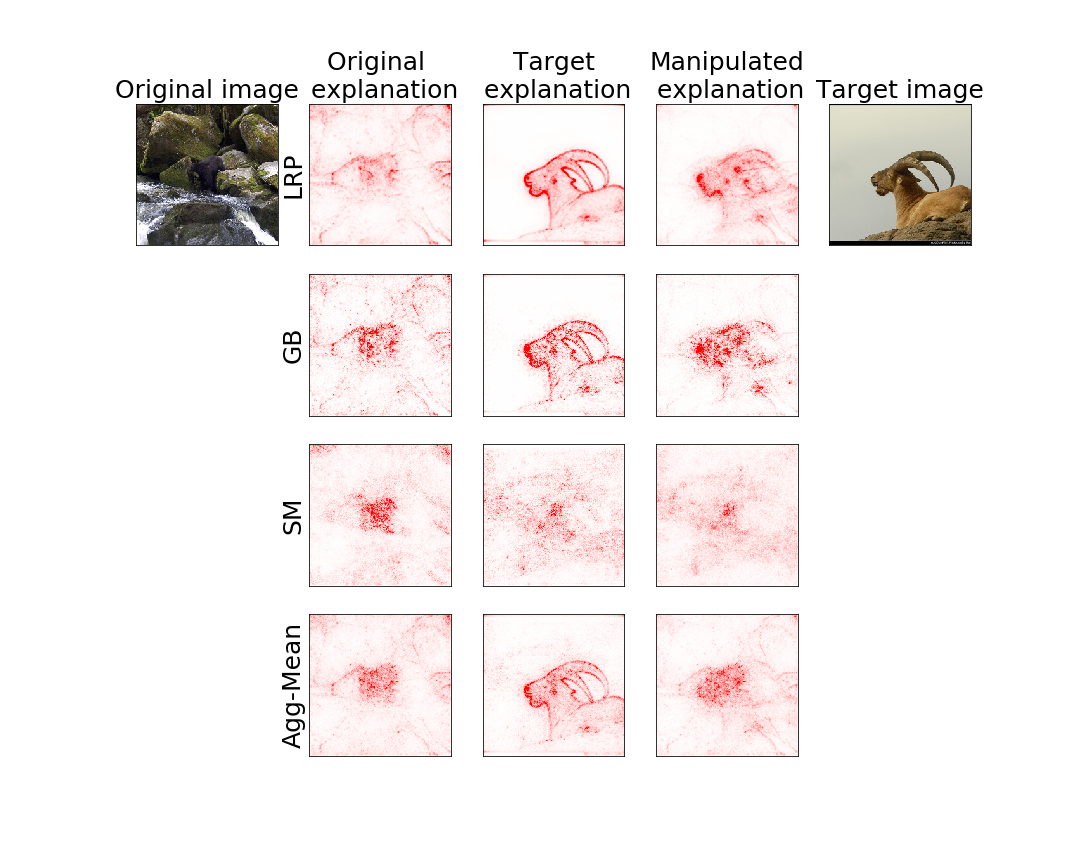}
	\end{center}
	\caption{Appendix example 3}
	\label{fig:attack2}
\end{figure*}

We provide more examples showing different explanation methods being attacked in \cref{fig:attack0,fig:attack1,fig:attack2,fig:attack3}. An abridged version of \cref{fig:attack0} is also shown in the main text.

\clearpage

\subsubsection{Transferability of attacks}
\begin{figure*}[hbt]
	\begin{center}
		\includegraphics[width=.5\linewidth]{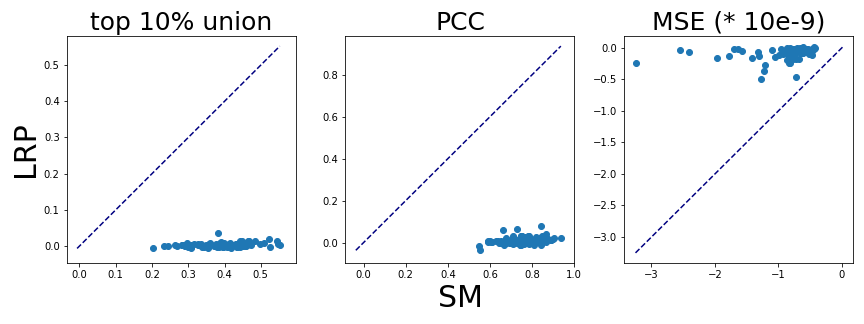}
		\includegraphics[width=.5\linewidth]{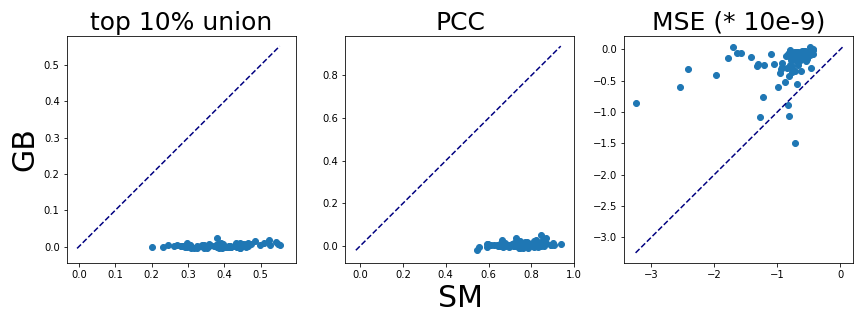}
		\includegraphics[width=.5\linewidth]{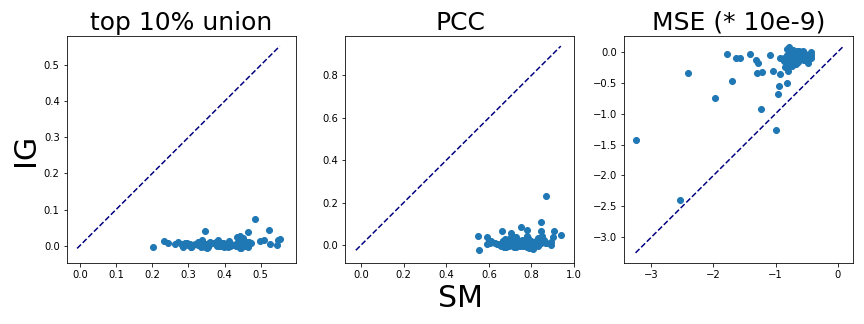}
	\end{center}
	\caption{Gradient as starting method}
	\label{fig:gradient_start}
\end{figure*}

\begin{figure*}[hbt]
	\begin{center}
		\includegraphics[width=.5\linewidth]{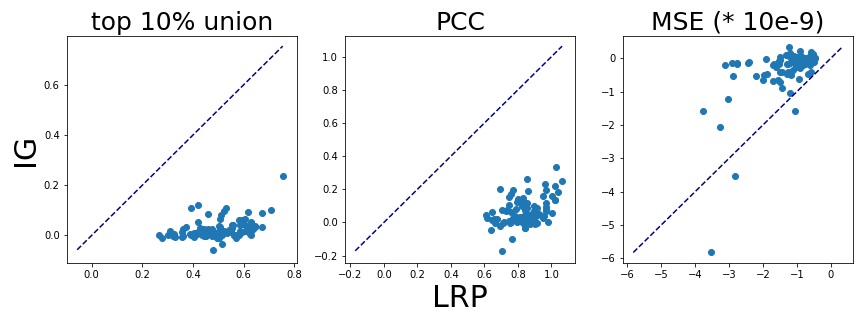}
		\includegraphics[width=.5\linewidth]{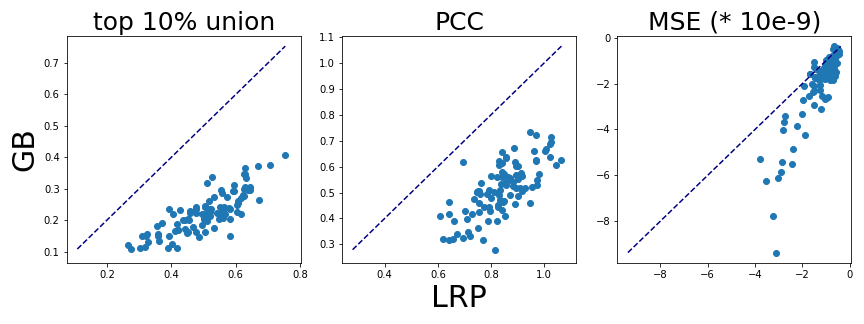}
		\includegraphics[width=.5\linewidth]{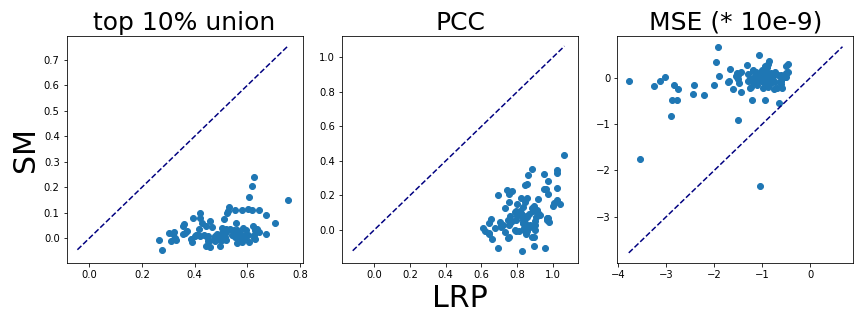}
	\end{center}
	\caption{LRP as starting method}
	\label{fig:lrp_start}
\end{figure*}

\begin{figure*}[hbt]
	\begin{center}
		\includegraphics[width=.5\linewidth]{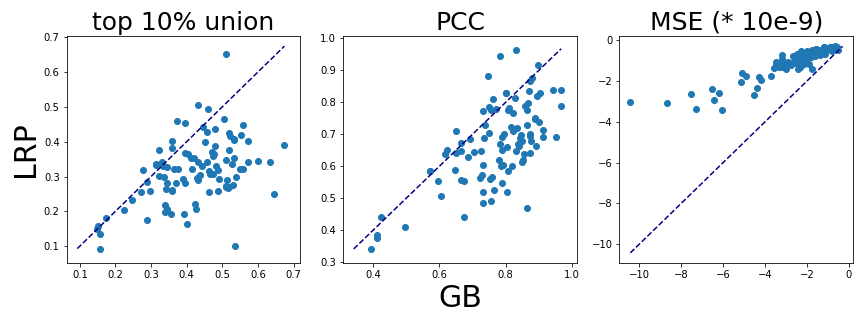}
		\includegraphics[width=.5\linewidth]{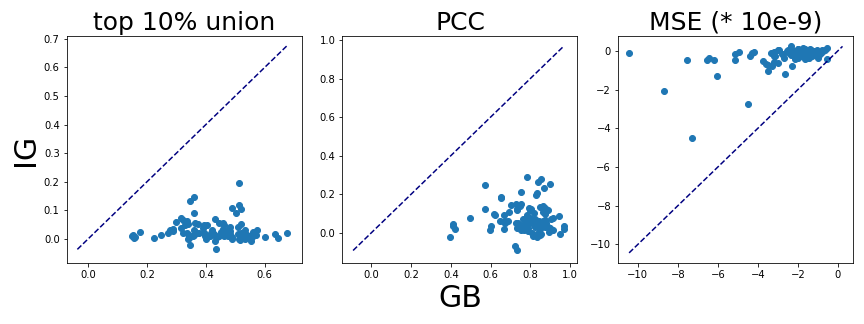}
		\includegraphics[width=.5\linewidth]{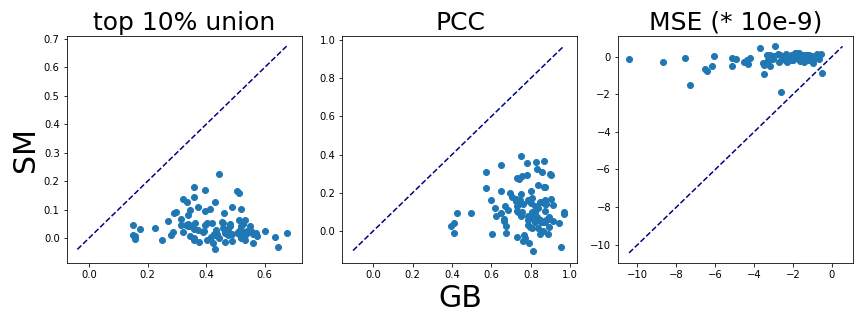}
	\end{center}
	\caption{GuidedBackprop as starting method}
	\label{fig:gb_start}
\end{figure*}

In the main text we show similarity metrics differences between the method being attacked and not being attacked for Guided Backprop and LRP. Here we provide scatter plots for the rest of the considered methods in \cref{fig:ig_start,fig:gradient_start,fig:lrp_start,fig:gb_start}:
\begin{figure*}[hbt]
	\begin{center}
		\includegraphics[width=.5\linewidth]{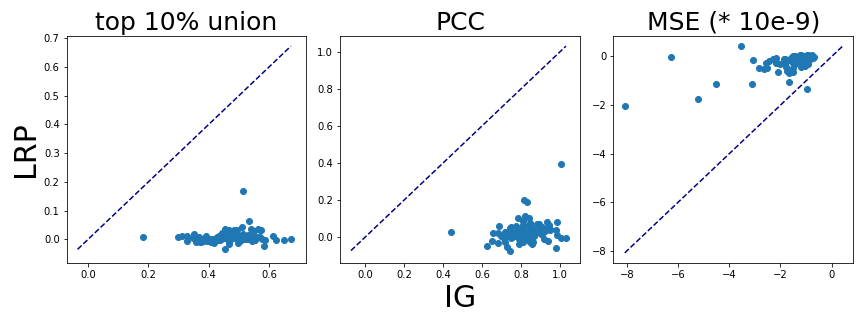}
		\includegraphics[width=.5\linewidth]{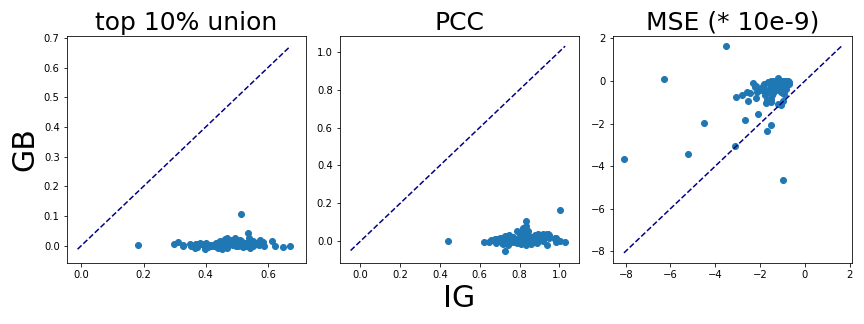}
		\includegraphics[width=.5\linewidth]{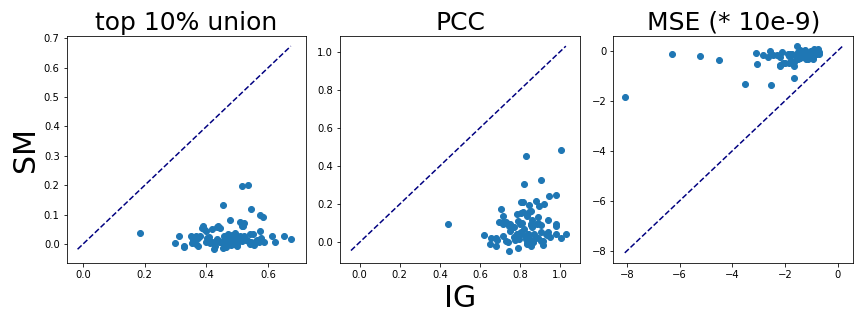}
	\end{center}
	\caption{Integrated Gradient as starting method}
	\label{fig:ig_start}
\end{figure*}

\clearpage
\subsubsection{Similarity of the attacked images to the starting images}
We provide the average distance of the adversarial images to the original images in \cref{tab:results_attacks_extended} (calculated in RGB space, average over all pixels). As can be seen  in \cref{fig:attack0,fig:attack1,fig:attack2,fig:attack3,}, there is no visual difference to the input images for any of the attacks. Attacking \textit{AGG-Mean} has the smallest distance to the input image, supporting our hypothesis that aggregating explanation methods removes vulnerabilities to adversarial manipulation.
\begin{table*}[!hbt]
	\caption{Evaluation scores across methods and architectures on a hundred samples, including similarity of the resulting image to the starting image. Deviation is SE. }
	\label{tab:results_attacks_extended}
	\vskip 0.15in
	\begin{center}
		\begin{small}
			\begin{sc}

				\begin{tabular}{lrrrr}
					\toprule
					{} & MSE $ \Delta $ (*10e-9) & PCC & top 10\% Union &MSE on images \\
					method & & & \\
					\midrule
					SM 		& -0.92 	$\pm$ 0.00				& 0.74 $\pm$ 0.01 			& 0.40 $\pm$ 0.01 & 0.0027 $\pm$ 0.0002\\
					GB 				& -3.25	 $\pm$ 0.02			& 0.77 $\pm$ 0.01			& 0.42 $\pm$ 0.01 &0.0110 $\pm$ 0.0025\\
					LRP & -1.45 $\pm$ 0.01						& 0.81 $\pm$ 0.01 			& 0.49 $\pm$ 0.01  &0.0047 $\pm$ 0.0006\\
					IG 				& -1.76 $\pm$ 0.01			& 0.82 $\pm$ 0.01			& 0.47 $\pm$ 0.01 & 0.0102 $\pm$ 0.0022\\
					AGG-Mean 	& \textbf{-0.89} $\pm$ 0.01 	& \textbf{ 0.54 $\pm$ 0.01 }	& \textbf{ 0.24 $\pm$ 0.01} & 0.0013 $\pm$ 0.0001 \\
					\bottomrule
				\end{tabular}
				
			\end{sc}
		\end{small}
	\end{center}
	\vskip -0.15in
\end{table*}
\clearpage
\subsubsection{Other attacks}
In the main text we mainly concern ourselves with making the explanation of one image look like a pre-specified target explanation, as this is a use case where the motivation of an attacker is apparent. However, as introduced in \citep{Ghorbani2019} other attack objectives are also conceivable. 

\begin{figure*}[hbt]
	\begin{center}
		\includegraphics[width=.7\linewidth]{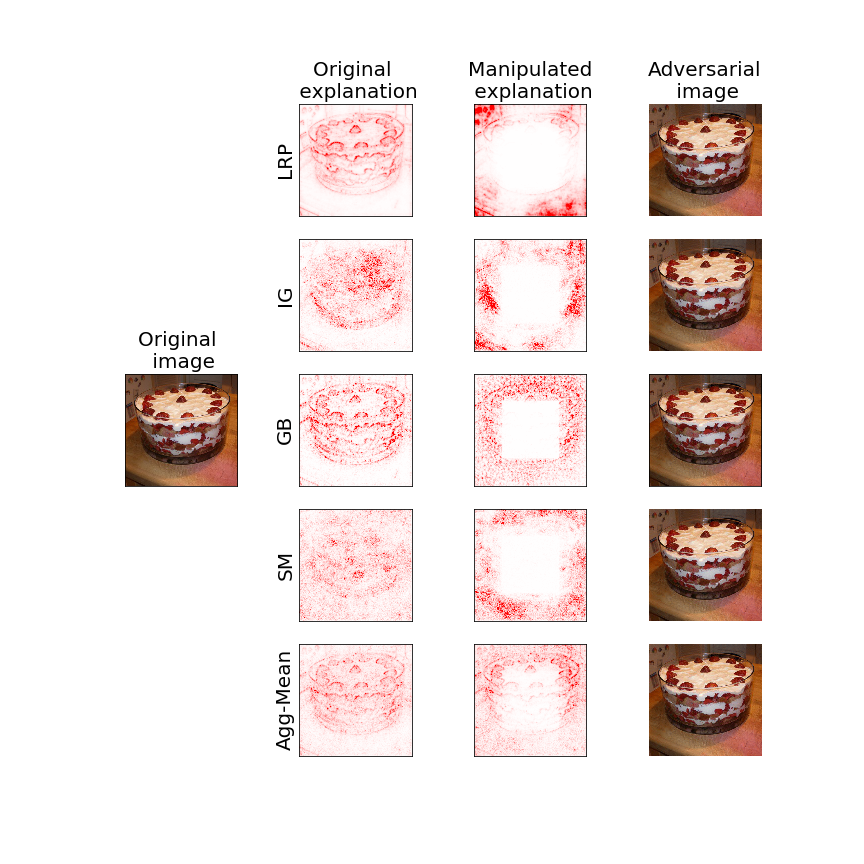}
	\end{center}
	\caption{Attacking explanation methods to make an area irrelevant as in \citep{Ghorbani2019}. AGG-Mean is most robust.}
	\label{fig:square_attack0}
\end{figure*}

\begin{figure*}[hbt]
	\begin{center}
		\includegraphics[width=.7\linewidth]{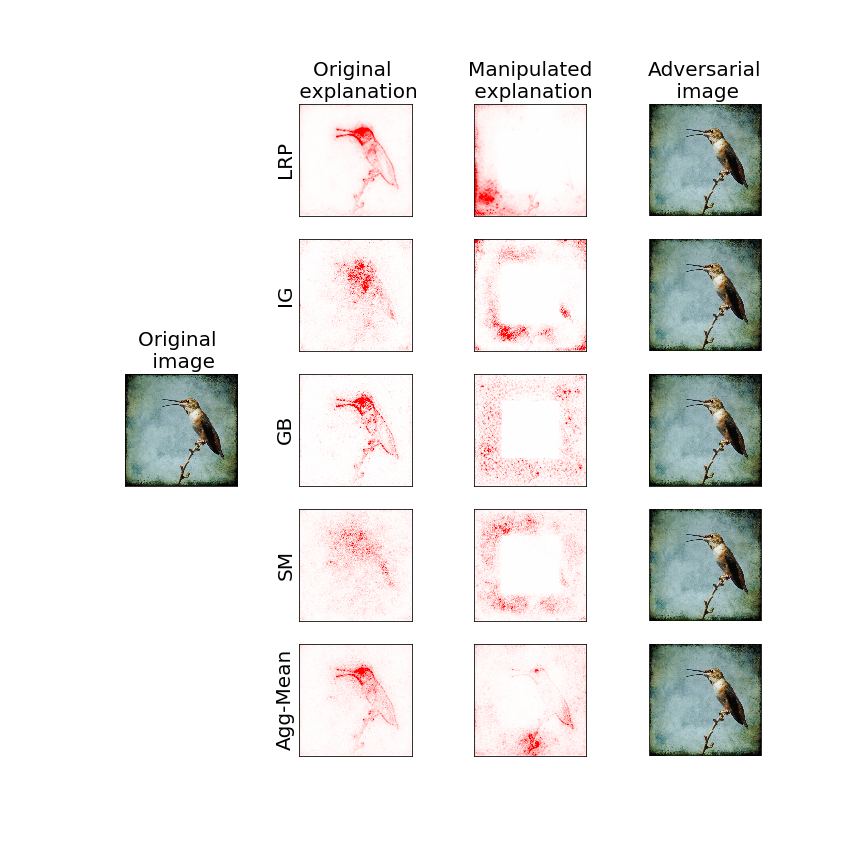}
	\end{center}
	\caption{Attacking explanation methods to make an area irrelevant as in \citep{Ghorbani2019}. AGG-Mean is most robust.}
	\label{fig:square_attack1}
\end{figure*}

We show results when following the objective of making a specified area of the explanation not relevant, i.e.\ a blank space in the explanation as introduced in \citep{Ghorbani2019}. A square (in size a quarter of the image) centered on the middle should not contain any relevance for the explanation. Size and position of the blank space were chosen ad-hoc, we assume that the center of the image generally contains useful information for the classification. We show quantitative results in \cref{tab:square_res}, computing how much percentage of the original relevance is preserved and qualitative results in \cref{fig:square_attack0,fig:square_attack1}. While an aggregation is not completely robust to the attack, far more of the original explanation is preserved. 

\begin{table*}[!hbt]
	\caption{Manipulating explanations to show a blank (irrelevant) square. Aggregating explanation methods preserves far more of the original explanation than any single method.  }
	\label{tab:square_res}
	\vskip 0.15in
	\begin{center}
		\begin{small}
			\begin{sc}
				
				\begin{tabular}{lrrr}
					\toprule
					{} &  Start &  After attack &  Percentage \\
					method          &                   &                 &            \\
					\midrule
					SM        &              0.34 &        2.05e-02 &       0.06 \\
					GB &              0.41 &        7.57e-03 &       0.02 \\
					IG &              0.38 &        1.48e-02 &       0.04 \\
					LRP             &              0.37 &        4.11e-03 &       0.01 \\
					AGG-Mean     &              0.37 &        \textbf{5.10e-02} &      \textbf{ 0.14} \\
					\bottomrule
				\end{tabular}
				
			\end{sc}
		\end{small}
	\end{center}
	\vskip -0.15in
\end{table*}

\clearpage

\end{document}